\documentclass[10pt,twocolumn,letterpaper]{article}

\usepackage[pagenumbers]{cvpr}      
\definecolor{cvprblue}{rgb}{0.21,0.49,0.74}
\usepackage[pagebackref,breaklinks,colorlinks,allcolors=cvprblue]{hyperref}

\usepackage{booktabs}
\usepackage{multirow}
\usepackage{threeparttable}

\def\paperID{} 
\def\confName{CVPR}
\def\confYear{2026}

\title{Model Agnostic Preference Optimization for Medical Image Segmentation}

\author{
Yunseong Nam$^1$ \quad
Jiwon Jang$^1$ \quad
Dongkyu Won$^2$\quad
Sang Hyun Park$^2$ \quad
Soopil Kim$^3$  \\ \\
$^1$DGIST, School of Undergrad Studies, Daegu, Republic of Korea \\
$^2$DGIST, Department of Robotics and Mechatronics Engineering, Daegu, Republic of Korea \\
$^3$DGIST, Division of Intelligent Robot, Daegu, Republic of Korea \\
{\tt\small \{nys3015, soopilkim\}@dgist.ac.kr}
}

\begin{document}
\maketitle
\begin{abstract}
Preference optimization offers a scalable supervision paradigm based on relative preference signals, yet prior attempts in medical image segmentation remain model-specific and rely on low-diversity prediction sampling. In this paper, we propose MAPO (Model-Agnostic Preference Optimization), a training framework that utilizes Dropout-driven stochastic segmentation hypotheses to construct preference-consistent gradients without direct ground-truth supervision. MAPO is fully architecture- and dimensionality-agnostic, supporting 2D/3D CNN and Transformer-based segmentation pipelines. Comprehensive evaluations across diverse medical datasets reveal that MAPO consistently enhances boundary adherence, reduces overfitting, and yields more stable optimization dynamics compared to conventional supervised training.
\end{abstract}    
\section{Introduction}
\label{sec:introduction}
Medical image segmentation is a core task in medical image analysis \citep{gao2025medical}, for which numerous deep learning–based methods have been proposed \citep{ronneberger2015unet, Zhou2021ModelsGenesis, yu2019uamt}. Recent progress has been largely driven by Foundation Models (FMs) trained on large-scale medical imaging datasets \citep{kirillov2023segment}, Large Language Models (LLMs) trained on medical reports \citep{urooj2025survey}, and multimodal fusion frameworks that jointly leverage both modalities \citep{zhao2025sat,tang2025causal}. 
These models achieve impressive segmentation accuracy and generalization. However, their performance declines when a domain gap exists between training and clinical deployment due to differences in imaging protocols, modalities, target organs, or disease types \citep{zhang2020generalizing,guan2022domain}. Therefore, fine-tuning is often essential to ensure clinically reliable segmentation performance in real-world settings.

Human Preference Optimization (HPO) is one fine-tuning paradigm that guides model learning using prediction outcomes aligned with human or expert preferences \citep{wirth2017survey}. Initially introduced to align LLMs with user intentions, preference optimization has evolved through methods such as Reinforcement Learning from Human Feedback (RLHF) \citep{ouyang2022training}, Proximal Policy Optimization (PPO) \citep{schulman2017ppo}, and Direct Preference Optimization (DPO) \citep{rafailov2023dpo}. Among these, DPO has gained particular attention for its conceptual simplicity and efficiency. In this approach, a model generates multiple predictions for a single input, and a preference dataset is constructed based on human- or rule-defined good and bad examples (\ie preferred and less preferred outputs, respectively); the model is then optimized to increase similarity with good examples and decrease similarity with bad ones. Since the preference signal relies on relative comparison rather than absolute ground truth, this learning paradigm offers lower annotation costs, robustness to noise, and improved training stability and generalization \citep{jinnai2024aepo,wirth2017survey}.

Recently, preference optimization has been applied to prompt-based medical image segmentation models \citep{konwer2025enhancing,wu2025sampo}. However, existing studies face two key limitations. (1) They are restricted to specific 2D SAM-based architectures, making it difficult to generalize across diverse medical imaging domains—ranging from 2D to 3D and even 4D spatio-temporal modalities \citep{wu2025sampo,zhao2025sat}. (2) For deterministic segmentation models, the methods for obtaining diverse prediction candidates (e.g., thresholding) are insufficient to capture adequate prediction variability \citep{konwer2025enhancing}. As a result, meaningful good/bad examples cannot be effectively obtained, limiting the utility of preference-based training.
These challenges naturally lead to the following research question: Can preference optimization be effectively applied to general medical image segmentation models beyond foundation models?

To address this question and overcome identified limitations, we propose MAPO (Model-Agnostic Preference Optimization), a dropout-based preference optimization framework applicable to a wide range of medical image segmentation architectures. MAPO employs dropout to generate diverse and informative preference samples. Traditionally, dropout has been used for regularization and uncertainty estimation. In our approach, by varying the dropout rate, we enable the model to produce multiple diverse predictions for a single input, as different neurons are randomly deactivated. This creates a robust preference set for optimization. Since our method is model-agnostic, it can be seamlessly extended to 2D CNNs, Transformer-based networks, and 3D medical segmentation models. Furthermore, we adopt an online preference learning strategy, where the model iteratively regenerates preference sets and updates itself through preference optimization. This feedback loop progressively refines segmentation performance and stabilizes training.

To validate the effectiveness of our proposed method, we conducted extensive experiments across multiple medical imaging modalities, model architectures, and loss functions. The results demonstrate that MAPO achieves superior segmentation performance and stable training compared to existing approaches.
Our key contributions are summarized as follows:
\begin{itemize}
    \item We propose the first preference optimization framework applicable to medical image segmentation models regardless of architecture. By leveraging dropout, MAPO generates diverse and informative preference data that enable robust and consistent learning.
    \item The proposed framework eliminates the dependence on naïve and heuristic sampling strategies (e.g., thresholding) by utilizing dropout-based online training, thereby enabling progressive preference optimization with minimal human intervention.
    \item Through extensive experiments on various architectures and medical datasets, we verify that our proposed framework consistently delivers improved segmentation performance and stable optimization behavior compared to existing methods.
\end{itemize}

\section{Related Work}
\label{sec:relatedwork}

\subsection{Human Preference Optimization of Deep Neural Networks}
Human Preference Optimization (HPO) aims to align model outputs with human preferences by incorporating preference signals into the fine-tuning process. This paradigm has recently expanded beyond large language models (LLMs) to segmentation, speech recognition, and multimodal tasks. Early approaches were based on Reinforcement Learning from Human Feedback (RLHF). Ouyang et al.~\cite{ouyang2022instructgpt} demonstrated that RLHF, using reward model trained on human preferences and Proximal Policy Optimization (PPO)~\cite{schulman2017ppo}, can enhance a language model, improving the output quality over larger base models.

To simplify RLHF, subsequent works sought to remove the explicit reward model while still leveraging human feedback. Yuan et al.~\cite{yuan2023rrhf} proposed Rank Responses with Human Feedback (RRHF), which aligns a model to human preferences using a ranking loss over sampled responses rather than a separate reward model. Bai et al.~\cite{bai2022constitutional,bai2023rlaif} introduced Constitutional AI (CAI) and Reinforcement Learning from AI Feedback (RLAIF), which replaces human labels with model-generated critiques. These methods generate self-critiques and revisions, and followed by fine-tuning with AI-generated preferences to align the model.


More recently, direct optimization of preferences has emerged. Rafailov et al.~\cite{rafailov2023dpo} proposed Direct Preference Optimization (DPO), which removes both the reward model and the RL loop by directly optimizing a log-likelihood ratio so that preferred responses have higher probabilities than disfavored ones. Azar et al.~\cite{azar2023psipo} generalised RLHF and DPO under a unified $\Psi$PO objective and introduced Identity Preference Optimization (IPO) as a special case that optimizes directly over preferences without approximations. Ethayarajh et al.~\cite{ethayarajh2024kto} proposed KTO (Model Alignment as Prospect Theoretic Optimization), which treats human feedback through prospect theory and introduces a “human-aware” loss that maximizes perceived utility rather than log-likelihood. Hong et al.~\cite{hong2024orpo} further simplified preference alignment with ORPO (Odds Ratio Preference Optimization), a reference-model–free, monolithic algorithm that contrasts favored and disfavored responses via an odds ratio penalty.

Preference optimization has also been extended to image segmentation tasks. Konwer et al.~\cite{konwer2025enhancing} enhanced the Segment Anything Model (SAM) for medical imaging by combining unsupervised prompts with DPO-based alignment; their method uses ratings or rankings from a virtual annotator to guide segmentation. Wu et al.~\cite{wu2025sampo} proposed SAMPO (Segment Anything Model with Preference Optimization), which applies preference optimization to train foundation models for capturing high-level segmentation intent from sparse interactions. These approaches demonstrate that preference-based fine-tuning can improve segmentation quality, yet their reliance on specific backbone models (e.g., SAM) limits broader medical applicability. To overcome this, we propose a model-agnostic preference optimization framework applicable to a wide range of medical segmentation models, independent on specific model architectures.

\begin{figure*}[t!]
    \centering
    \includegraphics[width=1.0\textwidth]{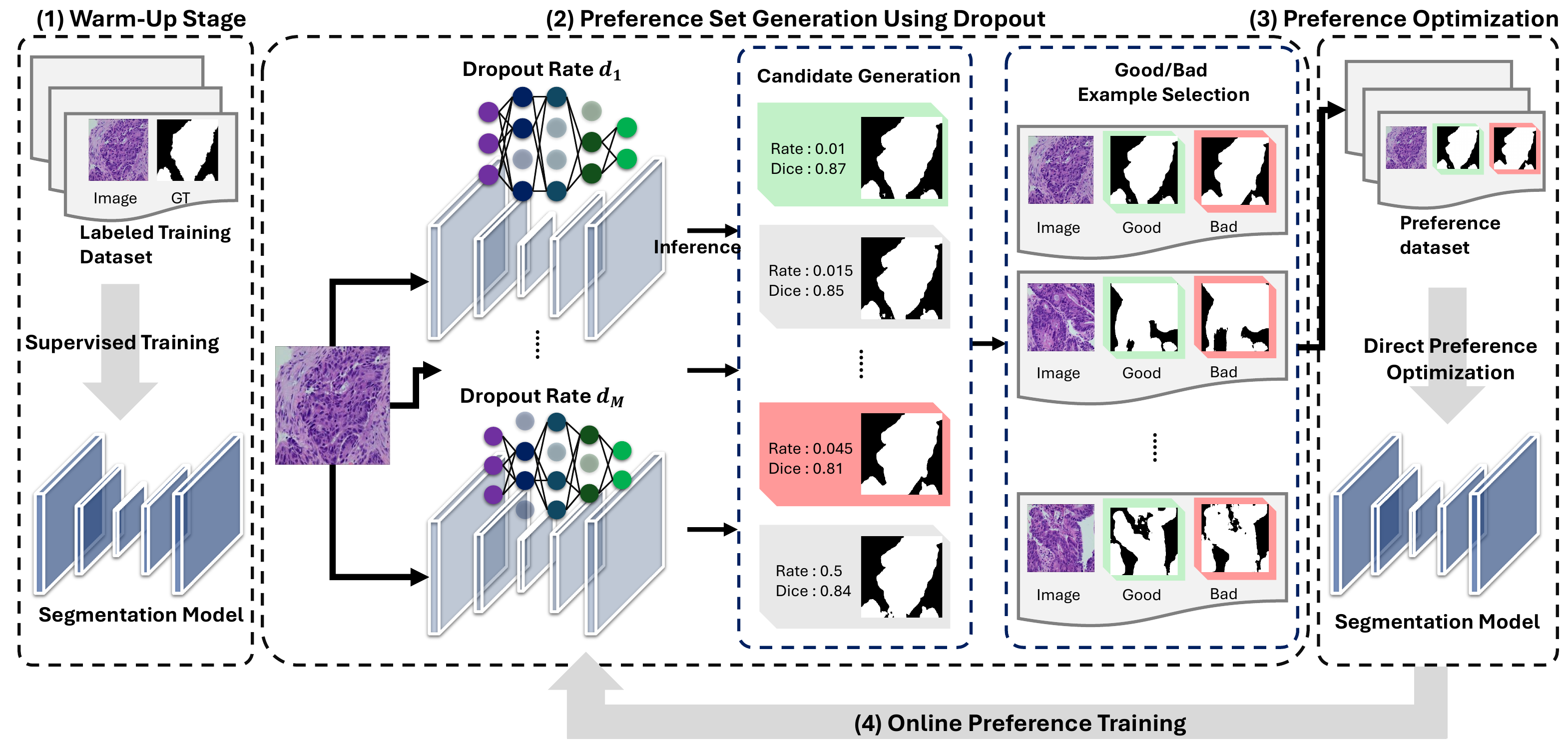}
    \caption{Overview of our proposed model-agnostic preference optimization for medical image segmentation, which consists of 4 stages: (1) Warm-up stage, (2) Preference set generation using dropout, (3) preference optimization, (4) Online preference training.}
    \label{fig:method}
\end{figure*}

\subsection{Medical Image Segmentation}
Deep learning–based medical image segmentation was initially driven by CNN encoder–decoder architectures, most notably U-Net and its variants \cite{ronneberger2015unet,milletari2016vnet}, which leverage skip connections to recover spatial details. These architectures were further extended to attention-enhanced models such as Attention U-Net \cite{Oktay2018AttentionUNet} and the highly automated and optimized nnU-Net framework \cite{Isensee2021nnUNet}.
More recently, Transformer-based and hybrid architectures—including TransUNet \cite{chen2021transunet}, Swin-Unet \cite{cao2021swinunet}, UNETR \cite{Hatamizadeh2022UNETR}, and Swin-UNETR \cite{Hatamizadeh2022SwinUNETR}—have been introduced to better capture both global contextual cues and local semantic details. By modeling long-range dependencies and complex spatial relationships, these models achieve improved accuracy in delineating intricate anatomical structures such as vascular networks, organs, and tumors.

Alongside architectural advances, various loss functions such as Dice \cite{milletari2016vnet}, Tversky \cite{Salehi2017Tversky}, Focal \cite{Lin2017Focal}, Boundary \cite{Kervadec2019Boundary}, and Lovász-Softmax \cite{Berman2018Lovasz}—have been introduced to mitigate class imbalance and boundary sensitivity, improving training stability and contour accuracy. However, these loss functions often rely on large labeled datasets and struggle to generalize across diverse modalities or small structures like micro-tumors, motivating the development of data-efficient learning strategies.

To address these challenges in data-scarce and diverse medical imaging scenarios, recent research has explored versatile learning paradigms for enhanced data efficiency and generalization. Meta-learning methods such as MAML \cite{Finn2017MAML} enable fast adaptation to new organs or domains with limited labels, while semi- and self-supervised approaches—including Mean Teacher \cite{Tarvainen2017MeanTeacher}, UA-MT \cite{yu2019uamt}, and SASSNet \cite{li2020sassnet}—leverage consistency regularization and pseudo-labeling to exploit unlabeled data. Self-supervised pretraining frameworks like Models Genesis \cite{Zhou2021ModelsGenesis} and MoCo \cite{He2020MoCo} further enhance representation quality. 

Building on these data-efficient learning strategies, foundation models have recently emerged as a unified paradigm for visual understanding across diverse domains. The Segment Anything Model (SAM) demonstrates strong zero-shot segmentation capability in natural images, yet its performance remains limited in medical imaging due to modality shifts and anatomical variability~\cite{mazurowski2023segment,chen2024segmentanything}. To address this gap, lightweight adaptation techniques have been explored, including parameter-efficient adapters~\cite{gao2025medical,nam2024instasam,Chen2023SAMAdapter}, test-time adaptation strategies such as SAM-TTA~\cite{wu2025sam}, and hierarchical decoders like H-SAM~\cite{cheng2024unleashing}, which enhance structural consistency without retraining the full model.

More recently, studies have shown that incorporating preference optimization during foundation-model adaptation can outperform conventional direct supervision, as it encourages models to align with clinically preferred outputs rather than solely matching ground-truth annotations~\cite{konwer2025enhancing,wu2025sampo}. Motivated by this finding, we extend the use of preference optimization beyond foundation-model adaptation and generalize it to the broader landscape of medical image segmentation. Specifically, we propose a preference optimization–based framework that can be seamlessly applied to diverse architectures—including 2D CNNs, transformer-based networks, and 3D segmentation models—allowing standard segmentation pipelines to benefit from preference-aligned learning.

\section{Method}
\label{sec:method}

\subsection{Overview}
We propose a novel preference-based training strategy that distinguishes between the model’s good and bad predictions (\ie preferred and less preferred model outputs, respectively), leveraging their differences as informative learning signals. This approach enables the model to learn from its own prediction quality, enhancing segmentation performance. Importantly, our method is architecture-agnostic and can be applied to any segmentation network.

The overall training pipeline consists of four stages: (1) warm-up training, (2) preference data generation, (3) preference optimization, and (4) online preference training. First, we train a segmentation model from scratch to establish a baseline level of performance. Then, we perform dropout-based stochastic inference to generate multiple prediction candidates for each image. By evaluating these candidates against the ground truth mask, we automatically select good and bad example pairs to construct a preference set. The model is further fine-tuned using these preference pairs with DPO objective, to increase the likelihood of good predictions while suppressing bad ones. 

The iterative approach begins once the model's performance improves. We regenerate updated preference data using the refined model and iterate the optimization process over multiple rounds. This progressive refinement allows the model to continually absorb preference information, yielding stronger generalization capability.

\subsection{Preliminary: Direct Preference Optimization}
\label{sec:dpo_prelim}

DPO \cite{rafailov2023dpo} is a recent preference alignment technique that directly optimizes a model using human preference signals without learning an explicit reward function. Unlike RLHF methods \cite{ouyang2022instructgpt}, which require training a reward model and performing reinforcement learning with Kullback-Leibler (KL) regularization, DPO simplifies preference learning by directly optimizing a likelihood-based objective.

Given an input $x$, the model produces two outputs: a preferred response $y_{+}$ and a less-preferred response $y_{-}$. Human preference can be expressed using a pairwise comparison prior, modeled as:
\begin{equation}
P(y_{+} > y_{-} \mid x) = 
\frac{\exp(r(x,y_{+}))}{
\exp(r(x,y_{+})) + \exp(r(x,y_{-}))},
\label{eq:bt_dpo}
\end{equation}
where $r(\cdot)$ denotes an explicit reward.

DPO bypasses explicit reward learning by modeling reward differences as a log-ratio between the optimized policy $\pi_\theta$ and the reference policy $\pi_{\mathrm{ref}}$:
\begin{equation}
\mathcal{L}_{\mathrm{DPO}} = - \log \sigma \left(\beta 
\log \frac{\pi_\theta(y_{+}\mid x)}{\pi_{\mathrm{ref}}(y_{+}\mid x)} - \beta \log \frac{\pi_\theta(y_{-}\mid x)}{\pi_{\mathrm{ref}}(y_{-}\mid x)} \right),
\label{eq:dpo_core}
\end{equation}
where $\sigma(\cdot)$ is the logistic function and $\beta$ adjusts the sensitivity to preference differences. Thus, DPO directly increases the likelihood of preferred samples while penalizing disfavored outcomes, functioning as an explicit reward model within a simple supervised learning setup. Due to its fully differentiable objective and absence of a separate reward model, DPO offers a lightweight yet effective alignment approach. This flexibility extends beyond language modeling to visual domains, such as image segmentation proposals, where preference feedback is critical.

\subsection{Model Agnostic Preference Optimization}
\subsubsection{Warm-up Stage}
In this stage, the model is trained using direct supervision, which is similar to conventional segmentation approaches by employing losses such as Cross-Entropy or Dice loss. To generate preference data, dropout is activated in each layer during training to produce diverse prediction candidates. Our baseline model is selected based on the best validation performance and used as the initial model for the next stage.

\subsubsection{Dropout-driven Preference Data Generation}
To construct preference data, we apply dropout during inference with a set of dropout rates $\mathcal{D} = \{d_1, d_2, \dots, d_M\}$ to generate diverse predictions. For each training sample, we generate ${K}$ prediction candidates $\{\hat{\mathbf{y}}_{k}\}_{k=1}^{K}$ each corresponding to a model output with a dropout rate $d \in \mathcal{D}$. We evaluate the quality of each prediction using the ground-truth segmentation mask $\mathbf{y}^{gt}$.

First, we select the index of the prediction with the highest Dice score as the good example:
\begin{equation}
k_{+} = \arg\max_{k} \text{Dice}(\hat{\mathbf{y}}_{k}, \mathbf{y}^{gt}), 
\qquad
\hat{\mathbf{y}}_{+} = \hat{\mathbf{y}}_{k_{+}}.
\label{eq:kplus}
\end{equation}
To ensure training stability, we avoid selecting bad examples with performance gaps that could destabilize optimization.
We define a valid candidate set $\mathcal{C}$ by selecting predictions where the Dice score gap from the good example exceeds a threshold $\tau$, ensuring meaningful differences:
\begin{equation}
\mathcal{C} = \left\{ k \mid
\text{Dice}(\hat{\mathbf{y}}_{+}, \mathbf{y}^{gt}) -
\text{Dice}(\hat{\mathbf{y}}_{k}, \mathbf{y}^{gt}) \ge \tau \right\}.
\end{equation}

Finally, we select the index of the worst prediction within this candidate set as the bad example:
\begin{equation}
k_{-} = \arg\min_{k \in \mathcal{C}} \text{Dice}(\hat{\mathbf{y}}_{k}, \mathbf{y}^{gt}),
\qquad
\hat{\mathbf{y}}_{-} = \hat{\mathbf{y}}_{k_{-}}.
\label{eq:kminus}
\end{equation}
This selection strategy enables the extraction of diverse yet meaningful preference pairs $(\hat{\mathbf{y}}_{+}, \hat{\mathbf{y}}_{-})$, while maintaining stable optimization behavior.

\subsubsection{Preference Optimization}
In this stage, the model is optimized using preference data without dropout operations to focus on preference data.
Specifically, given a preferred prediction $\hat{\mathbf{y}}_{+}$ and a less-preferred prediction $\hat{\mathbf{y}}_{-}$ from the same input $\mathbf{x}$, our objective is to explicitly increase the likelihood of $\hat{\mathbf{y}}_{+}$ while discouraging $\hat{\mathbf{y}}_{-}$. The segmentation model $\pi_\theta$ outputs voxel-wise prediction probabilities $p_\theta \in [0,1]^{|\Omega|}$, where $\Omega$ denotes the set of spatial indices.

We compute the likelihood of a prediction $\hat{\mathbf{y}}$ using a voxel-wise Bernoulli log-likelihood:
\begin{equation}
\log \pi_\theta(\hat{\mathbf{y}} \mid \mathbf{x})
=
\frac{1}{|\Omega|}
\sum_{i \in \Omega}
\left[
\hat{y}_{i} \log p_{\theta,i}
+
(1 - \hat{y}_{i}) \log (1 - p_{\theta,i})
\right].
\label{eq:likelihood_voxel_simplified}
\end{equation}
Following DPO, we enforce preference consistency by maximizing the log-likelihood ratio between a trainable model $\pi_\theta$ and a frozen reference model $\pi_{\text{ref}}$.

As noted in prior studies \cite{wu2025sampo}, optimizing a model with DPO loss alone often results in unstable training dynamics. To maintain segmentation accuracy while aligning preferences, we incorporate DPO with standard segmentation losses, ensuring stable and effective training. Our objective follows as:
\begin{equation}
\mathcal{L}
=
\lambda \big(
\mathcal{L}_{\text{Dice}}(p_\theta, \mathbf{y}^{gt})
+
\mathcal{L}_{\text{CE}}(p_\theta, \mathbf{y}^{gt})
\big)
+
\mathcal{L}_{\text{DPO}},
\label{eq:total_loss_simplified}
\end{equation}
where $\lambda$ denotes the weight factor for Dice and BCE terms. Although the model is supervised using ground-truth labels $\mathbf{y}^{gt}$, the inclusion of $\mathcal{L}_{\text{DPO}}$ enables exploration of a broader parameter space guided by preference consistency. This combination allows the model to achieve more robust and generalizable learning than relying solely on ground-truth supervision.

\begin{figure}[t!]
    \centering
    \includegraphics[width=1.0\columnwidth]{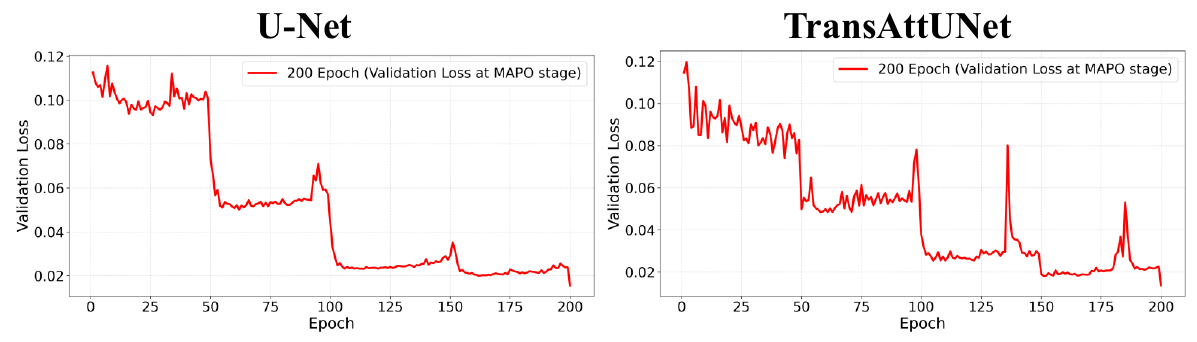}
    \caption{Validation loss ($\mathcal{L}_{CE}+\mathcal{L}_{Dice}$) \textit{v.s.} training epoch graph obtained in online preference training of U-Net \cite{ronneberger2015unet} and TransAttUNet \cite{chen2021transattunet} on EBHI dataset \cite{li2023ebhi}. They gradually decrease as we update the preference dataset every 50 epochs.}
    \label{fig:val_loss}
\end{figure}

\subsubsection{Online Preference Training}
Instead of using a static preference dataset, our framework dynamically updates preference pairs after each training round through iterative refinement. After each training round, we regenerate prediction candidates through dropout-based inference and update $\hat{\mathbf{y}}_{+}$ and $\hat{\mathbf{y}}_{-}$ using Eq.~(\ref{eq:kplus})--(\ref{eq:kminus}) to select optimal preference pairs, followed by further preference optimization.

This iterative preference update process ensures that the preference data aligns with the model’s evolving performance, allowing the preference signals to adapt to the model’s current limitations without requiring additional human intervention, as shown in Fig. \ref{fig:val_loss}. Consequently, the model enhances structural consistency and boundary precision, especially in challenging regions with ambiguous or uncertain segmentations, by leveraging dynamically updated preference data.

\section{Experiment}
\label{sec:experiment}

\begin{table*}[t]
\centering
\small
\begin{tabular}{l|l|l|l|l|l|l}
\hline
Backbone & Model & ISIC~\cite{tschandl2018isic} & EBHI~\cite{li2023ebhi} & FIVES~\cite{jin2022fives} & Kvasir~\cite{jha2020kvasirseg} & COVID-QU-Ex~\cite{chowdhury2021covidquex} \\ \hline
\multirow{6}{*}{CNN} 
 & U-Net~\cite{ronneberger2015unet} & 81.68 & 91.84 & 86.58 & 65.06 & 93.65 \\
 & U-Net~\cite{ronneberger2015unet}+ MAPO & 82.80(+1.12) & 91.96(+0.12) & 88.12(+1.54) & 74.08(+9.02) & 94.99(+1.34) \\
 & RollingUNet~\cite{liu2024rollingunet} & 83.32 & 86.05 & 80.46 & 90.78 & 91.32 \\
 & RollingUNet~\cite{liu2024rollingunet}+ MAPO & 82.97(-0.35) & 90.32(+4.27) & 83.06(+2.60) & 89.04(-1.74) & 96.53(+5.21) \\
 & U$^2$-Net~\cite{qin2020u2net} & 82.25 & 92.61 & 83.81 & 75.22 & 95.95 \\
 & U$^2$-Net~\cite{qin2020u2net}+ MAPO & 85.96(+3.71) & 92.64(+0.03) & 84.87(+1.06) & 84.09(+8.87) & 96.49(+0.54) \\ \hline
\multirow{4}{*}{ViT} 
 & SwinUNet~\cite{cao2021swinunet} & 78.65 & 90.75 & 82.79 & 52.20 & 92.61 \\
 & SwinUNet~\cite{cao2021swinunet}+ MAPO & 81.98(+3.33) & 92.25(+1.50) & 85.48(+3.69) & 59.83(+7.63) & 94.67(+2.06) \\
 & MISSFormer~\cite{huang2021missformer} & 84.04 & 91.45 & 73.77 & 58.03 & 93.93 \\
 & MISSFormer~\cite{huang2021missformer}+ MAPO & 85.87(+1.83) & 92.31(+0.86) & 74.09(+0.32) & 78.57(+20.54) & 95.12(+1.19) \\ \hline
\multirow{4}{*}{Hybrid} 
 & H2Former~\cite{he2023h2former} & 83.52 & 91.25 & 76.63 & 50.90 & 95.05 \\
 & H2Former~\cite{he2023h2former}+ MAPO & 84.99(+1.47) & 92.44(+1.19) & 82.17(+5.54) & 73.28(+22.38) & 96.05(+1.00) \\
 & TransAttUNet~\cite{chen2021transattunet} & 81.65 & 91.17 & 75.19 & 53.80 & 95.13 \\
 & TransAttUNet~\cite{chen2021transattunet}+ MAPO & 84.94(+3.39) & 92.95(+1.88) & 75.99(+0.80) & 81.73(+28.93) & 96.30(+1.27) \\ \hline
 \end{tabular}
\caption{Dice similarity scores (\%) of our proposed methods against baseline models using five 2D medical image segmentation datasets and 7 different methods. }
\label{tab:dice_results}
\end{table*}

\begin{table*}[t]
\centering
\small
\begin{tabular}{l|l|l|l|l|l|l}
\hline
Backbone & Model & ISIC~\cite{tschandl2018isic} & EBHI~\cite{li2023ebhi} & FIVES~\cite{jin2022fives} & Kvasir~\cite{jha2020kvasirseg} & COVID-QU-Ex~\cite{chowdhury2021covidquex} \\ \hline
\multirow{6}{*}{CNN} 
 & U-Net~\cite{ronneberger2015unet} & 5.33 & 2.01 & 6.25 & 36.60 & 1.39 \\
 & U-Net~\cite{ronneberger2015unet}+ MAPO & 4.51(-0.82) & 1.85(-0.16) & 6.17(-0.08) & 28.78(-7.82) & 1.20(-0.19) \\
 & RollingUNet~\cite{liu2024rollingunet} & 5.24 & 1.99 & 38.02 & 9.36 & 1.12 \\
 & RollingUNet~\cite{liu2024rollingunet}+ MAPO & 5.09(-0.15) & 1.53(-0.46) & 19.40(-18.62) & 11.48(+2.12) & 0.74(-0.38) \\
 & U$^2$-Net~\cite{qin2020u2net} & 5.09 & 1.26 & 19.87 & 24.00 & 0.81 \\
 & U$^2$-Net~\cite{qin2020u2net}+ MAPO & 4.12(-0.97) & 1.18(-0.08) & 20.00(+0.13) & 15.80(-8.20) & 0.69(+0.12) \\ \hline
\multirow{4}{*}{ViT} 
 & SwinUNet~\cite{cao2021swinunet} & 6.55 & 1.84 & 7.79 & 45.80 & 1.74 \\
 & SwinUNet~\cite{cao2021swinunet}+ MAPO & 5.78(-0.77) & 1.33(-0.51) & 6.74(-1.05) & 42.54(-3.26) & 1.14(-6.00) \\
 & MISSFormer~\cite{huang2021missformer} & 3.73 & 1.93 & 9.24 & 40.47 & 1.66 \\
 & MISSFormer~\cite{huang2021missformer}+ MAPO & 3.78(+0.05) & 1.31(-0.62) & 9.07(-0.17) & 23.05(-17.42) & 1.07(-0.59) \\ \hline
\multirow{4}{*}{Hybrid} 
 & H2Former~\cite{he2023h2former} & 4.70 & 1.57 & 15.21 & 45.73 & 1.03 \\
 & H2Former~\cite{he2023h2former}+ MAPO & 4.06(-0.64) & 1.22(-0.35) & 5.55(-9.66) & 29.07(-16.66) & 0.83(-0.20) \\
 & TransAttUNet~\cite{chen2021transattunet} & 5.53 & 1.71 & 14.38 & 31.88 & 1.25 \\
 & TransAttUNet~\cite{chen2021transattunet}+ MAPO & 4.02(-1.51) & 1.30(-0.41) & 12.46(-1.92) & 22.58(-9.30) & 0.88(-0.37) \\ \hline
 \end{tabular}
\caption{ASD scores (\%) of our proposed methods against baseline models using five 2D medical image segmentation datasets and 7 different methods.}
\label{tab:asd_results}
\end{table*}

\begin{table*}[t]
\centering
\small
\begin{tabular}{l|l|ll|ll|ll}
\hline
 &  & \multicolumn{2}{c|}{Parse2022 \cite{Parse2022_dataset}} & \multicolumn{2}{c|}{Sliver07 \cite{SLIVER07}} & \multicolumn{2}{c}{Luna16 \cite{LUNA16}} \\ \cline{3-8} 
Backbone & Model & Dice & ASD & Dice & ASD & Dice & ASD \\ \hline
\multirow{4}{*}{CNN} 
& VNet \cite{milletari2016vnet} & 66.25 & 13.20 & 71.23 & 30.79 & 96.98 & 2.18 \\
& VNet \cite{milletari2016vnet}+MAPO & 68.79(+2.54) & 12.90(-0.30) & 73.40(+2.17) & 27.94(-2.85) & 96.99(+0.01) & 1.56(-0.62) \\
& DynUNet \cite{Hadlich2023DynUNet} & 72.95 & 17.78 & 62.90 & 27.69 & 95.96 & 5.10 \\
& DynUNet \cite{Hadlich2023DynUNet}+MAPO & 74.20(+1.25) & 15.20(-2.58) & 65.25(+2.35) & 21.22(-6.47) & 94.65(-1.31) & 4.90(-0.20) \\ \hline

\multirow{2}{*}{ViT} 
& DFormer \cite{Wu2023DFormer} & 65.57 & 15.71 & 67.60 & 26.87 & 96.98 & 3.93 \\
& DFormer \cite{Wu2023DFormer}+MAPO & 67.60(+2.03) & 13.11(-2.60) & 69.50(+1.90) & 21.11(-5.76) & 97.02(+0.04) & 3.90(-0.03) \\ \hline

\multirow{2}{*}{Hybrid} 
& UNETR \cite{Hatamizadeh2022UNETR} & 60.01 & 18.15 & 72.31 & 26.27 & 96.47 & 1.32 \\
& UNETR \cite{Hatamizadeh2022UNETR}+MAPO & 62.76(+2.75) & 15.47(-2.68) & 74.93(+2.62) & 21.35(-4.92) & 97.01(+0.54) & 4.26(+2.94) \\ \hline
\end{tabular}
\caption{Dice similarity and ASD scores (\%) of our proposed methods against baseline models using three 3D medical image segmentation datasets and 4 different methods.}
\label{tab:3d_results}
\end{table*}

\subsection{Experimental Setting}
To validate the effectiveness of the proposed DPO-based preference alignment learning, we conducted extensive experiments on a wide range of medical image segmentation datasets. The 2D experiments were performed on five datasets: ISIC-2018 \cite{tschandl2018isic}, Kvasir-SEG \cite{jha2020kvasirseg}, FIVES \cite{jin2022fives}, EBHI-SEG \cite{li2023ebhi}, and COVID-QU-Ex \cite{chowdhury2021covidquex}, while the 3D experiments were conducted on Parse2022 \cite{Parse2022_dataset}, Sliver07 \cite{SLIVER07}, Luna16 \cite{LUNA16} datasets for binary segmentation evaluation. Detailed information about each dataset is provided in the appendix. These datasets differ in lesion morphology, size, contrast, and noise characteristics, making them well-suited for assessing whether the proposed approach consistently improves performance across diverse conditions. For evaluation, we used widely used metrics: the Dice Score and Average Surface Distance (ASD). All comparison methods were trained for the same number of epochs using MAPO-applied data, and the best models were selected based on validation performance.

\subsection{Implementation Details}
All models were implemented in PyTorch and trained on an NVIDIA RTX A5000 GPU with 24 GB of memory. The Adam optimizer was used for 2D experiments, while the AdamW optimizer was used for 3D experiments, with a fixed learning rate of $1\times10^{-4}$. For 2D data, all images were resized to $384\times384$, whereas for 3D data, the input volume sizes were adjusted depending on each dataset.
The 2D model was initially trained for 100 epochs using supervised learning, after which preference sets were generated every 50 epochs during an additional 200 epochs of DPO training, resulting in a total of 300 epochs. The 3D model followed the same strategy with 500 epochs of supervised pretraining and 800 epochs of DPO training, generating preference sets every 200 epochs for a total of 1300 epochs. 

To enhance prediction diversity during preference generation, dropout rates were sampled from 0.0 to 0.05 with 0.005 intervals for 2D models, and from 0.0 to 0.30 with 0.05 intervals for 3D models.
This online training procedure enables the model to initially learn pixel-level correspondence and later produce structurally more accurate segmentation results beyond pixel-wise accuracy. The hyperparameters were set to $\beta=0.1$, $\tau=0.3$, and $\lambda=0.5$.

\subsection{Results of Preference Optimization Application on Various Model Architectures}
The quantitative results summarized in Tables~\ref{tab:dice_results} and~\ref{tab:asd_results} collectively demonstrate that the proposed framework enhances segmentation performance across various architectures and datasets. This improvement is reflected in higher Dice similarity and reduced boundary errors, as measured the ASD, indicating that our framework consistently produces more anatomically coherent and perceptually refined segmentations.

To be specific, across the five datasets, we observed significant variability in the performance of different architectures—CNN-based, ViTs-based, and hybrid models, indicating that no single backbone dominates in all segmentation tasks. However, when our proposed method was applied, most models exhibited consistent performance improvements, with Dice similarity gains ranging from 1–2\% to over 5\% depending on the dataset and architecture.
For example, U$^2$-Net + MAPO increased Dice score by +3.71\% on ISIC and +8.87\% on Kvasir, while TransAttUNet + MAPO achieved +3.39\% improvement on ISIC. These results confirm that our proposed method provides a robust performance enhancements across diverse architectures.

Importantly, even when improvements in Dice score are modest, we frequently observe meaningful reductions in ASD, reflecting smoother boundaries and improved structural precision. For example, in RollingUNet on the ISIC dataset, the Dice score slightly decreases (-0.35\%), yet ASD still improves (-0.15), indicating better contour alignment despite minimal change in overlap. A similar trend appears in MISSFormer on EBHI, where the Dice score increases (+0.86\%) and ASD decreases (-0.62), suggesting that MAPO enhances boundary accuracy even when Dice performance is already nearing saturation.

We observed interesting results on Kvasir dataset: several Transformer-based models exhibited unstable baseline training behavior on certain datasets, resulting in notably low performance compared to other datasets. In these cases, MAPO played a stabilizing optimization, leading to substantial performance improvements. For example, TransAttUNet + MAPO boosted the Dice score by +28.93\% and reduced ASD by -9.30 on Kvasir, indicating that MAPO effectively mitigates optimization instability, enabling reliable reliable and structurally consistent segmentation outcomes.

Table~\ref{tab:3d_results} demonstrates consistent trends in 3D medical image segmentation tasks. Across diverse architectures (CNNs, Transformers, and hybrid models), MAPO consistently yields performance gains in most cases, confirming its applicability beyond 2D scenarios. For example, on the Sliver07 dataset, MAPO increases VNet's performance from 71.23 to 73.40 (+2.17\%) and reduces ASD from 30.79 to 27.94 (-2.85), suggesting that MAPO stabilizes optimization and improves boundary precision. Meanwhile, improvements on the LUNA16 dataset are relatively modest, as the task is less challenging and baseline models are near saturation, limiting further improvements.

\begin{figure}[t!]
    \centering
    \includegraphics[width=0.9\columnwidth]{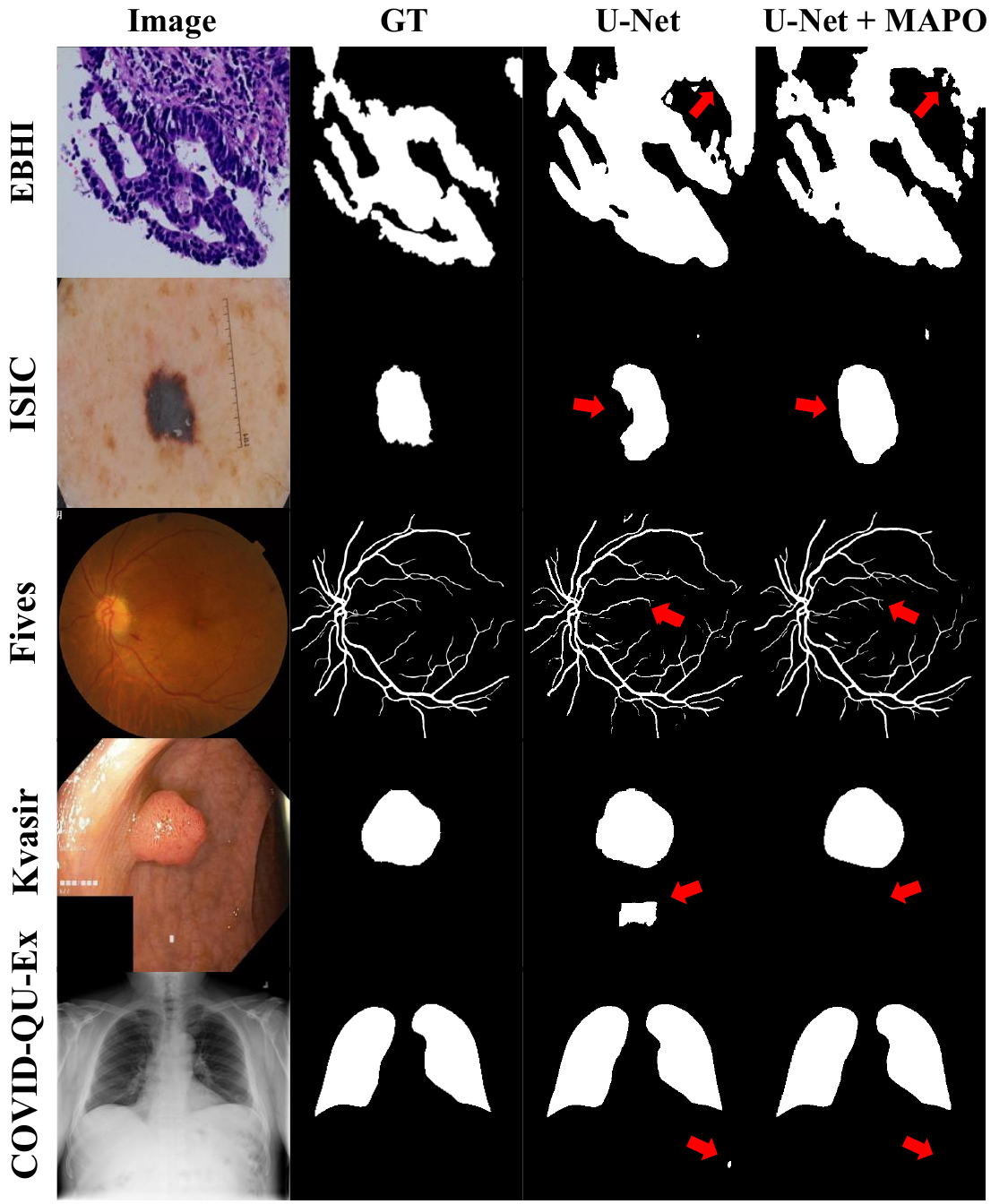}
    \caption{Qualitative comparison between our proposed method and baseline using U-Net \cite{ronneberger2015unet} architecture.}
    \label{fig:qual}
\end{figure}

Figure \ref{fig:qual} presents qualitative comparisons using U-Net across the five datasets, showing that MAPO-enhanced models consistently produce fewer false positives and more refined contours than their baseline counterparts. These results collectively show that MAPO generalizes well across diverse architectures and datasets.


\begin{figure}[t!]
    \centering
    \includegraphics[width=1.0\columnwidth]{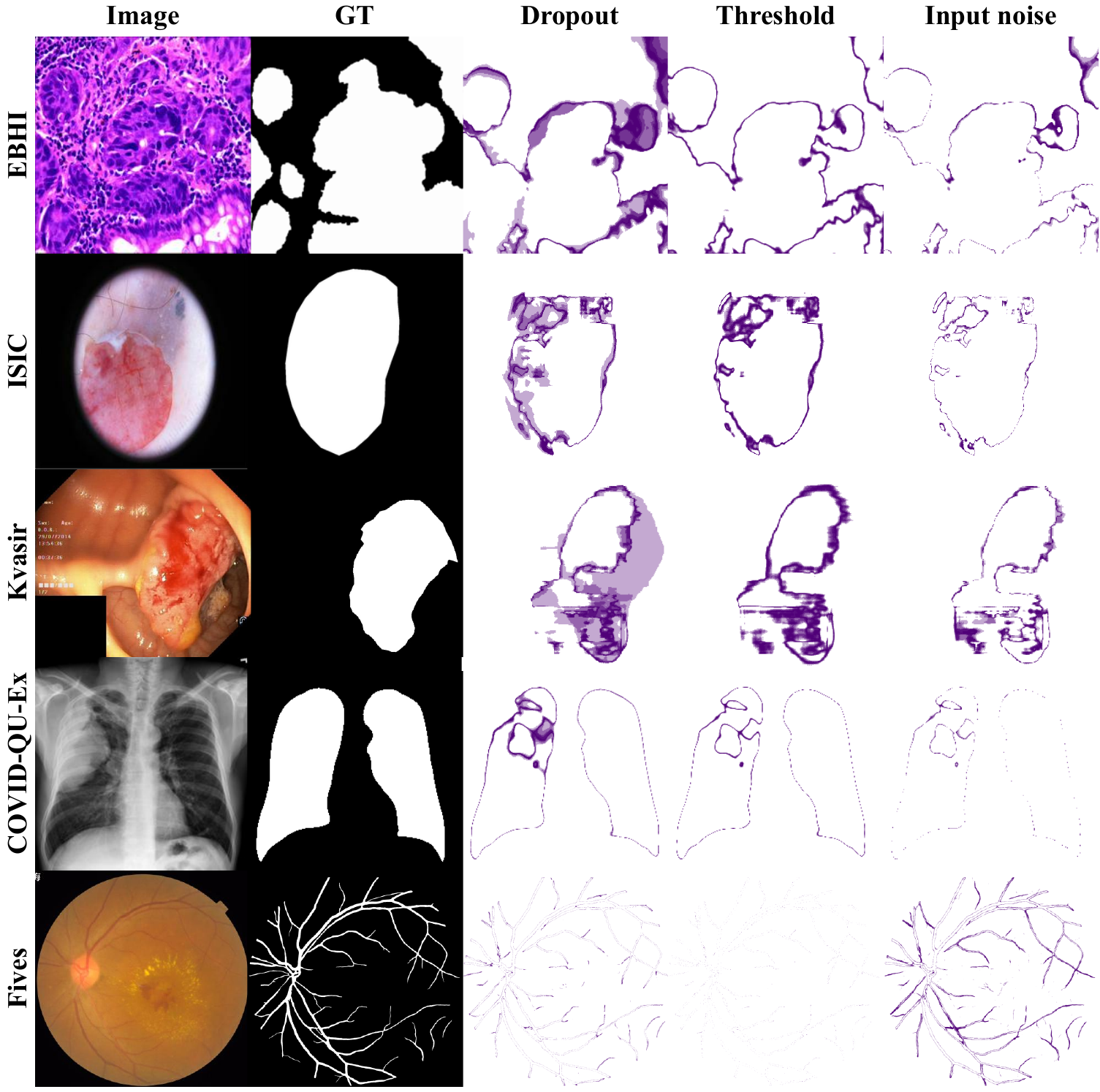}
    \caption{Visualization of pixel-level variance using different stochastic sampling strategies with U-Net \cite{ronneberger2015unet}.}
    \label{fig:qual_var}
\end{figure}

\begin{table}[t]
\centering
\small
\resizebox{\columnwidth}{!}{
\begin{tabular}{ll|cc|cc|cc}
\hline
 & \multicolumn{1}{l}{} & \multicolumn{2}{|c|}{ISIC \cite{tschandl2018isic}} & \multicolumn{2}{c|}{FIVES \cite{jin2022fives}} & \multicolumn{2}{c}{Kvasir \cite{jha2020kvasirseg}} \\ \cline{3-8} 
 & \multicolumn{1}{l|}{Method} & Dice & ASD & Dice & ASD & Dice & ASD \\ \hline

\multirow{4}{*}{Dice} 
& Baseline & 81.2 & 6.0 & 85.3 & 7.5 & 50.0 & 59.0 \\
& Dropout & 83.5 & 5.4 & 87.2 & 6.4 & 68.8 & 38.3 \\
& Threshold \cite{konwer2025enhancing} & 82.6 & 5.8 & 86.4 & 6.9 & 66.7 & 39.1 \\
& Input noise & 82.1 & 6.3 & 87.1 & 6.7 & 67.8 & 38.9 \\ \hline

\multirow{4}{*}{DiceCE} 
& Baseline & 81.7 & 5.3 & 86.6 & 6.3 & 65.1 & 36.6 \\
& Dropout & 82.8 & 4.5 & 88.1 & 6.2 & 74.1 & 28.8 \\
& Threshold \cite{konwer2025enhancing} & 82.8 & 5.7 & 87.5 & 6.4 & 68.4 & 36.4 \\
& Input noise & 82.6 & 6.6 & 88.0 & 6.1 & 68.3 & 37.9 \\ \hline

\end{tabular}}
\caption{Comparison of preference data generation strategies using U-Net \cite{ronneberger2015unet} architecture.}
\label{tab:abl_random}
\end{table}

\begin{table}[t]
\centering
\small
\resizebox{\columnwidth}{!}{
\begin{tabular}{l|c|c|c}
\hline
Loss function & RollingUNet \cite{liu2024rollingunet} & SwinUNet \cite{cao2021swinunet} & TransattUNet\cite{chen2021transattunet} \\ \hline
CE\cite{mao2023crossentropy} & 86.97 & 90.27 & 88.40 \\
Focal\cite{lin2017focalloss} & 86.45 & 78.39 & 88.27 \\
DiceCE\cite{goodfellow2016deep} & 87.01 & 90.75 & 91.17 \\
DiceFocal\cite{lin2017focalloss} & 86.81 & 91.11 & 91.10 \\
Dice\cite{milletari2016vnet} & 87.79 & 78.39 & 89.38 \\
Lovasz\cite{Berman2018Lovasz} & 86.74 & 91.02 & 88.79 \\
HD \cite{karimi2019hausdorff,celaya2023surface} & 83.17 & 83.11 & 82.18 \\
MAPO (Ours) & 90.32 & 92.25 & 92.95 \\ \hline
\end{tabular}}
\caption{Dice similarity scores of RollingUNet \cite{liu2024rollingunet}, SwinUNet \cite{cao2021swinunet}, and TransattUNet\cite{chen2021transattunet} using various loss functions on EBHI dataset \cite{li2023ebhi}.}
\label{tab:abl_loss}
\end{table}

\begin{table}[t]
\centering
\small
\begin{tabular}{c|cc|cc|cc}
\hline
 & \multicolumn{2}{c|}{ISIC \cite{tschandl2018isic}} & \multicolumn{2}{c|}{EBHI \cite{li2023ebhi}} & \multicolumn{2}{c}{Kvasir \cite{jha2020kvasirseg}} \\
$\tau$ & Dice & ASD & Dice & ASD & Dice & ASD \\ \hline 
0.5 & 80.45 & 6.76 & 91.85 & 1.83 & 67.09 & 35.74 \\
0.4 & 81.08 & 6.71 & 91.53 & 2.21 & 72.26 & 33.99 \\
0.3 & 82.80 & 4.51 & 91.96 & 1.85 & 74.08 & 28.78 \\
0.2 & 80.92 & 7.14 & 91.78 & 1.89 & 73.75 & 29.13 \\
0.1 & 81.26 & 6.70 & 91.15 & 1.93 & 73.89 & 27.06 \\ \hline
\end{tabular}
\caption{Ablation study on $\tau$ value.}
\label{tab:tau}
\end{table}



\subsection{Ablation Study on Various Stochastic Sampling Strategies}
To investigate how different preference set constructions affect DPO-based optimization, we compared three stochastic perturbation strategies—dropout-based sampling, threshold variation, and input noise injection—using a U-Net segmentation model. As shown in Table \ref{tab:abl_random}, the dropout-based approach consistently achieved the highest Dice and lowest ASD scores.

The threshold-based preference sets were generated by applying 11 thresholds ranging from 0.375 to 0.625 at 0.025 intervals to the model’s predicted segmentation maps, while the noise-based preference sets were created by adding Gaussian noise to the input images with standard deviations from 0 to 0.05 at 0.005 intervals. Note that all methods are trained from the same baseline warm-up model for fairness.

Among the three strategies, dropout-based sampling proves the most effective, as it naturally generates diverse segmentation predictions through random neurons deactivation during inference. This stochasticity generates preference pairs that effectively capture structural uncertainties in medical imaging.
Fig.~\ref{fig:qual_var} visualizes the pixel-wise variance in predictions generated by each strategy, with higher intensity indicating regions of greater prediction fluctuation.

Under threshold variation and input noise, differences between good and bad predictions were subtle, resulting in limited preference diversity. For both approaches, most variations were confined to boundary regions, indicating that the generated candidates lacked meaningful structural diversity. In contrast, our dropout strategy generated variance patterns across a broader spatial extent, showing that the model explores a richer set of plausible predictions and reveals a broader range of potential failure modes.

The FIVES dataset exhibited a slightly different trend. For input noise, many vascular regions showed consistently high variance, indicating instability rather than true diverse predictions. Moreover, threshold fails to generate vascular regions showing no significant regions for preference set. On the other hand, our dropout-based sampling produced a biologically coherent pattern: low variance in large, easily recognizable retinal vessels and high variance primarily in fine microvascular structures where segmentation is inherently challenging.

Overall, dropout-based preference generation provided richer and semantically more meaningful supervision, ultimately leading to improved overlap accuracy and smoother boundary delineation across datasets.

\subsection{Ablation on Loss Functions}
Furthermore, we compared the performance of existing loss functions in different architecture against the proposed method. Table~\ref{tab:abl_loss} reports the results obtained by training the baseline model using various loss functions. The results clearly show that the proposed method consistently outperforms the other loss functions.

Additionally, Table~\ref{tab:tau} reports the performance of the U-Net+MAPO model when varying $\tau$, which represents the Dice difference between the good and bad examples in different preference set generation. We observed that when the gap between good and bad predictions was either too large or too small, the model performance degraded. In contrast, providing a moderate and well-balanced difference led to improved performance. Based on these observations, we selected $\tau = 0.3$ for the final model.


\section{Conclusion}
\label{sec:conclusion}

In this work, we proposed MAPO (Model-Agnostic Preference Optimization), a general fine-tuning framework that extends preference optimization beyond foundation models to conventional medical image segmentation architectures. MAPO constructs reliable preference pairs by leveraging dropout-induced prediction diversity rather than architecture-dependent modules or heuristic sampling, enabling fully automated and supervision-free preference generation.
Across diverse datasets, architectures (CNNs, Transformers, and 3D models), and loss functions, MAPO consistently improved both segmentation accuracy and training stability compared to threshold- or noise-based strategies. These results demonstrate that properly formulated preference optimization offers a robust and generalizable mechanism for enhancing the reliability and alignment of medical image segmentation systems.

{
    \small
    \bibliographystyle{ieeenat_fullname}
    \bibliography{main}
}


\end{document}